\documentclass{article}

\usepackage[sglblindworkshop]{neurips_2025}

\usepackage[utf8]{inputenc} 
\usepackage[T1]{fontenc}    
\usepackage{hyperref}       
\usepackage{url}            
\usepackage{booktabs}       
\usepackage{amsfonts}       
\usepackage{nicefrac}       
\usepackage{microtype}      
\usepackage{xcolor}         
\usepackage{graphicx}
\usepackage{float}

\title{Enhancing Demand-Oriented Regionalization with Agentic AI and Local Heterogeneous Data for Adaptation Planning}
\workshoptitle{UrbanAI}
\author{
Seyedeh Mobina Noorani$^{1,2}$\thanks{Correspondence: s.noorani@ufl.edu},  
Shangde Gao$^{1,3}$, 
Changjie Chen$^{1,3}$, 
Karla Salda\~{n}a Ochoa$^{1}$\\[1em]
$^1$College of Design, Construction and Planning, University of Florida\\
$^2$Department of Electrical and Computer Engineering, University of Florida\\
$^3$Florida Institute for Built Environment Resilience (FIBER), University of Florida\\
Gainesville, FL 32611, USA
}
\begin{document}
\maketitle

\begin{abstract}
  Conventional planning units or urban regions, such as census tracts, zip codes, or neighborhoods, often do not capture the specific demands of local communities and lack the flexibility to implement effective strategies for hazard prevention or response. To support the creation of dynamic planning units, we introduce a planning support system with agentic AI that enables users to generate demand-oriented regions for disaster planning, integrating the human-in-the-loop principle for transparency and adaptability. The platform is built on a representative initialized spatially constrained self-organizing map (RepSC-SOM), extending traditional SOM with adaptive geographic filtering and region-growing refinement, while AI agents can reason, plan, and act to guide the process by suggesting input features, guiding spatial constraints, and supporting interactive exploration. We demonstrate the capabilities of the platform through a case study on the flooding-related risk 
  in Jacksonville, Florida, showing how it allows users to explore, generate, and evaluate regionalization interactively, combining computational rigor with user-driven decision making.
\end{abstract}


\section{Introduction}
Today's urban governance relies on planning units (or regions) to design and implement policies for growth, management, and adaptation to socioeconomic and environmental changes \cite{montero_delineation_2021}. 
These spatial frameworks allow planners to evaluate the distribution of resources \cite{us_epa_regionalization_2013}, assess policy impacts \cite{olagunju_integration_2016}, and ensure policy coherence on multiple scales \cite{dhar_multi-scale_2017}. 
The planning units also provide shared reference points to facilitate stakeholder participation and transparent decision making \cite{legacy_investigating_2010,matern_smart_2020}.

Existing planning units, such as administrative boundaries or census divisions, were designed for specific purposes, such as to speed up mail delivery, and were often misaligned with the demands of urban governance.  For example, census units can reflect population-based data collection rather than hazard exposure or social vulnerability, while other regions may not capture spatial patterns of climate risk \cite{nowak_spatial_2023}.  The rigidity of these units, combined with limited local knowledge and inconsistent fine-grained data, limits the ability of planners to design regions tailored to specific disaster or adaptation objectives \cite{heikkinen_fine-grained_2020,roest_mapping_2023}.

Data-driven regionalization can offer a promising solution by integrating multidimensional local data to produce homogeneous spatially contiguous regions aligned with planning objectives \cite{aydin_quantitative_2021,kueppers_data-driven_2020,zhang_advancing_2024}. 
Unlike conventional units, data-driven regionalization has the potential to capture social, environmental and physical heterogeneity and can support evidence-based decision making for targeted interventions. 
However, implementing such regionalizations remains challenging due to the complexity of constraints, the need for interpretability, and the iterative nature of planning decisions.

Recent advances in large language models (LLMs) and agentic AI provide new opportunities to support planners who lack training in geospatial or geocoding analysis in this process \cite{fu_large_2025,zhou_large_2024}. 
By combining generative reasoning, constraint-aware guidance, and human-in-the-loop interaction, agentic AI can help planners develop spatial divisions and iteratively refine generated regions \cite{feng_citygpt_2025, fu_large_2025,zhou_large_2024}. 
Such platforms can integrate multi-dimensional data, offer interpretable outputs, and facilitate collaborative exploration, transforming regionalization into an adaptive, interactive process aligned with local priorities and disaster-specific needs.

Building on this opportunity, we present an agentic AI based planning support system that implements a data-driven demand-oriented regionalization framework, RepSC-SOM (\underline{Rep}resentative-initialized, \underline{S}patially \underline{C}onstrained \underline{S}elf-\underline{O}rganizing \underline{M}ap). 
The core regionalization framework extends traditional SOM with representative-based initialization, adaptive geographic filtering, and region-growing refinement, while AI agents can assist users in suggesting input features, guiding spatial constraints, and supporting interactive exploration. 
The system can produce spatially coherent regions that capture localized vulnerabilities and risk profiles of hazards by integrating high-quality local socioeconomic and environmental data.
Through interactive exploration and human-in-the-loop refinement, the platform can enhance flexibility, transparency, and adaptability in spatial planning, supporting precise resource allocation, prioritization of interventions, and coordinated responses in disaster risk management and climate adaptation.

\section{Methods}
\subsection{Architecture of Agentic AI-Enhanced Planning Support System for Regionalization}
\begin{figure}[h]
    \centering
    \includegraphics[width=0.65\textwidth]{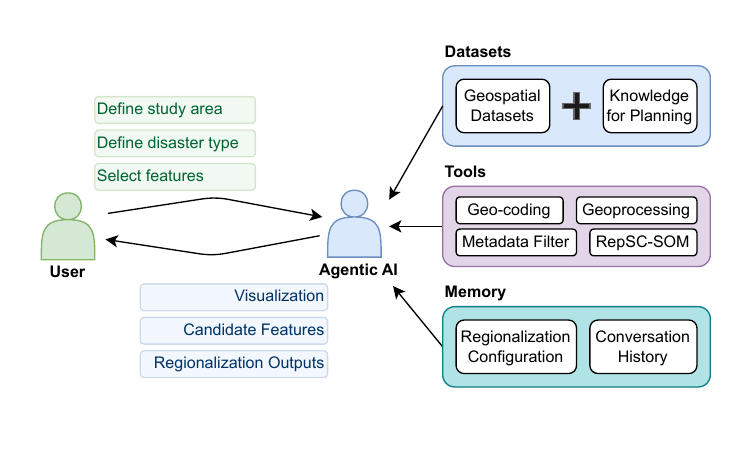}
    \caption{Architecture of the Agentic AI-enhanced planning support system for regionalization}
    \label{Figure0}
\end{figure}

We present a central planning agent to orchestrate the regionalization workflow (Fig.~\ref{Figure0}). This agentic AI conversational system for hazard-aware regionalization turns open-ended planning questions into defensible end-to-end analyses. The system is designed to facilitate transparent, adaptive, and user-driven spatial analysis through natural language interaction. Users engage with the platform through a conversational interface, specifying their study area and hazard of interest. The agentic back-end interprets these inputs and autonomously coordinates a sequence of analytical steps: geocoding the user’s location, selecting and configuring relevant geospatial datasets (e.g., Florida Geographic Data Library \cite{uf_geoplan_center_florida_2025}), and visualizing spatial features as interactive map layers. The agent further summarizes the characteristics of the data set and obtains user preferences for feature selection and the number of regions to generate. Once the user’s preferences are set, the agent calls the RepSC-SOM regionalization algorithm, integrating the resulting spatial partitions into the map interface for user exploration. Throughout the process, the agent maintains the context dialogue and adapts to user responses, while ensuring a seamless experience. Therefore, this architecture enables non-technical users to perform advanced spatial analyses and explore alternative scenarios, lowering barriers to the application of machine learning in geospatial decision support.

\subsection{Overview of RepSC-SOM Framework for Demand-Oriented Data-Driven Regionalization}
In the RepSC-SOM framework, regionalization is carried out through a SOM-based three-step \textbf{\textit{Embedding-Clustering-Refining}} process, in which the study area is represented as a grid cell raster. 
The outputs at each step are transparent and interpretable, enabling users to adjust the configurations and intervene in the process as needed. 
In the \textbf{\textit{Embedding}} step, the input features selected by users are projected into higher dimensional latent spaces using an auto-encoder \cite{li_comprehensive_2023}, which can capture the complex interactions and dependencies among the input features. 
Meanwhile, the SOM is initialized according to the geographic threshold, which is based on the semivariogram of the data of input features and is the basis for both the number and the initial states of the neurons. 
The \textbf{\textit{Clustering}} step then uses the \textbf{\textit{Embedding}} output as the input, iteratively assigning grid cells in the study area to their Best Matching Units (BMUs) and updating neurons with their spatial neighbors, thereby forming preliminary clusters.
In this process, the selection of the BMU for each grid cell involves two stages: candidate SOM neurons are first filtered within a geographic threshold using the Haversine distance \cite{maria_measure_2020}, and then the most similar SOM neuron in the feature space is chosen as the BMU. 
The weight of each SOM neuron is updated based on the features of the grid cells that selected it as their BMU. 
After multiple iterations, the SOM assigns each grid cell the weight of its BMU. 
Finally, in the \textbf{\textit{Refining}} step, the spatially constrained SOM output is postprocessed to improve spatial compactness and reduce fragmentation: grid cells are first spatially partitioned into initial regions, which are then iteratively merged through a region-growing process guided by the number of regions expected by users, the feature similarity, and spatial constraints. 
The output of the \textbf{\textit{Refining}} step is the regionalization result and is then presented to the users.

\section{Demonstration of the Proposed Agentic AI-Enhanced Planning Support System.}
We demonstrate the capabilities of our proposed system through case studies in urban areas facing diverse climate and disaster risks. The demonstration highlights how the proposed platform supports planners in designing targeted intervention strategies to address hazard-specific vulnerabilities by integrating local data, identifying priority areas, and refining regional boundaries.

\begin{figure}[H]
    \centering
    \includegraphics[width=0.9\textwidth]{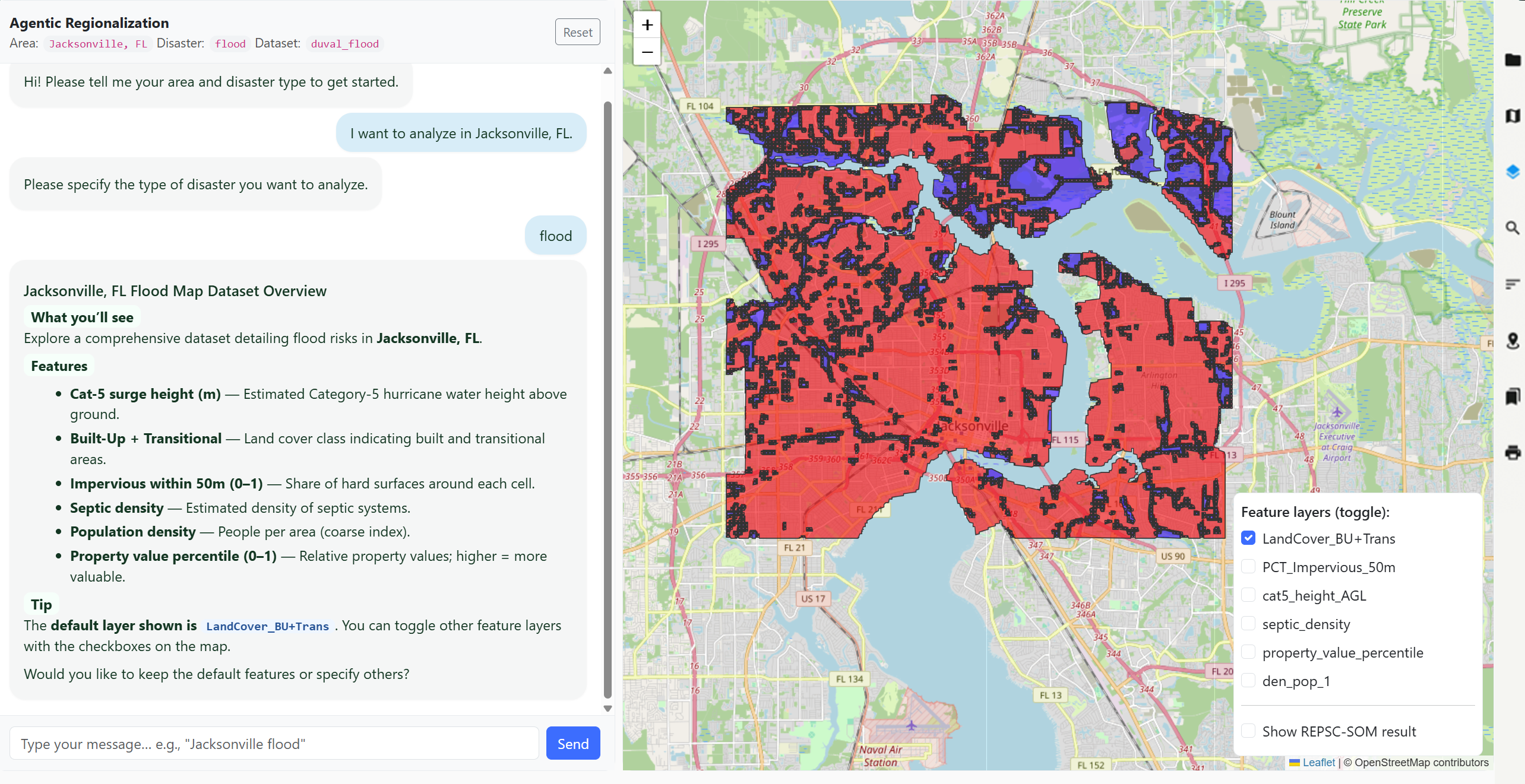}
    \caption{Screenshot of the proposed platform showing input features for regionalization based on user-specified study area and disaster type.}
    \label{Figure1}
\end{figure}

\textbf{Interactive Selection of Input Features Based on Study Area and Disaster Type.}
Users can begin by specifying their study area (Jacksonville, FL, in this case) and the type of disaster (flooding, in this case) through the platform interface. 
Based on the selected region and the disaster type, the LLM-based system dynamically selects and presents the candidate features for regionalization, 
such as socioeconomic indicators, environmental conditions, and the infrastructure vulnerability (Fig.~\ref{Figure1}).

\begin{figure}[H]
    \centering
    \includegraphics[width=0.9\textwidth]{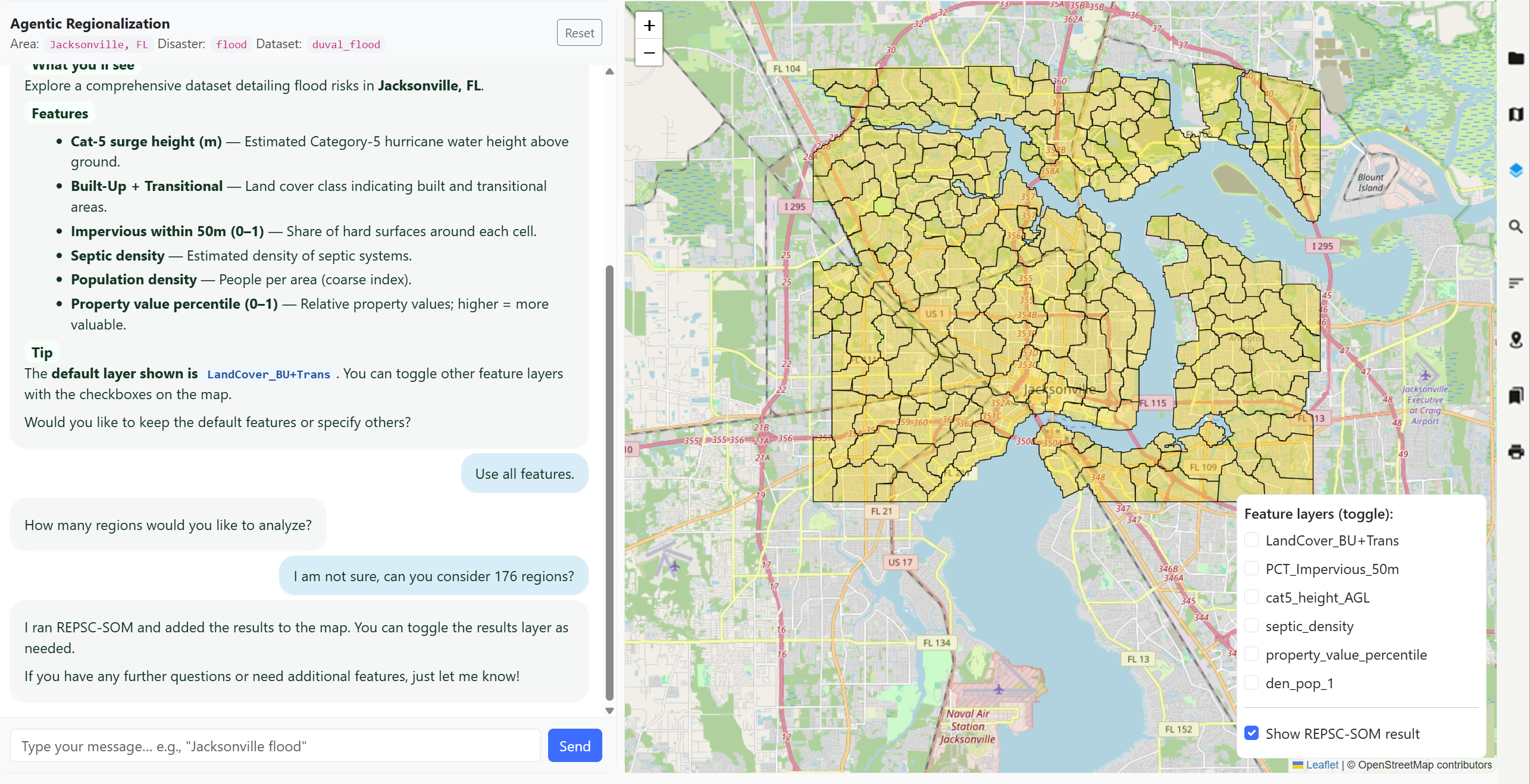}
    \caption{Screenshot of the proposed platform displaying the regionalization output.}
    \label{Figure2}
\end{figure}

\textbf{Regionalization Using the RepSC-SOM Framework.}
After selecting the suggested features, users specify the desired number of regions. The system then runs the RepSC-SOM regionalization and presents the results on an interactive map (Fig.~\ref{Figure2}), where users can interpret and evaluate the output.

\textbf{Human-in-the-Loop Refinement.}
The platform supports an iterative human-in-the-loop workflow, allowing users to provide feedback, adjust input features, or explore alternative configurations based on the regionalization output.
AI agents can respond to user input by updating the regionalization, ensuring that the final regions reflect both computational rigor and user knowledge. 
This process enables planners to generate demand-oriented, adaptive planning regions tailored to local priorities and disaster-specific considerations.

\section{Conclusion}
Our study demonstrates that a planning support system with agentic AI 
can support urban planners without a coding background to generate planning units that better capture local disaster risks than conventional units. 
By operationalizing the RepSC-SOM framework and leveraging heterogeneous local socioeconomic and environmental data, the platform can address mismatches between traditional boundaries and spatial heterogeneity, allowing the identification of priority areas, optimized resource allocation, and evidence-based adaptive urban governance. 
Its interactive, human-in-the-loop design allows regions to flexibly respond to dynamic hazard patterns and evolving urban conditions, supporting resilient and equitable adaptation strategies. 

\section{Acknowledgment}
This work was supported by the University of Florida (UF) AI and Complex Computational Research Award. The authors thank UFIT Research Computing and NVIDIA’s NVAITC at UF for providing computational resources and support that contributed to the research results reported in this publication. Any opinions, findings, conclusions, or recommendations expressed in this material are those of the authors and do not necessarily reflect the views of UF.

\bibliographystyle{plain}
\bibliography{NeurIPS}

\end{document}